\documentclass{article}
\usepackage{spconf,amsmath,amssymb,epsfig}
\usepackage{bm}
\usepackage{multirow}
\usepackage{rotating}

\renewcommand{\vec}[1]{\mbox{\boldmath$#1$}}

\graphicspath{
{figures/}
}

\title{Multiplicative updates for non-negative kernel SVM}
\name{Vamsi~K. Potluru\sthanks{Thanks to NIBIB grants 1 R01 EB 000840 and
    1R01 EB 005846}$^2$, Sergey~M. Plis\sthanks{Thanks to NIMH grant 1 R01
    MH076282-01.}$^2$,  Morten   M{\o}rup$^3$,  Vince~D.  Calhoun$^{1,2,4}$,
  Terran Lane$^2$} \address{
  $^1$Electrical and Computer Engineering Department, University  of New Mexico,New Mexico, USA, 87131\\
  $^2$Computer Science Department, University of
  New Mexico,
  New Mexico, USA, 87131\\
  $^3$DTU Informatics, Technical University of Denmark, Denmark\\
  $^4$Mind Research Network, New Mexico, USA, 87131}
\begin{document}  
%
\maketitle
\begin{abstract}
  We present  multiplicative updates for solving hard  and soft margin
  support  vector  machines  (SVM)  with non-negative  kernels.   They
  follow as a natural extension of the updates for non-negative matrix
  factorization.   No additional parameter  setting, such  as choosing
  learning,   rate   is   required.  Experiments   demonstrate   rapid
  convergence to good classifiers.  We analyze the rates of asymptotic
  convergence of the updates and  establish tight bounds.  We test the
  performance on  several datasets using  various non-negative kernels
  and report  equivalent generalization errors  to that of  a standard
  SVM.
\end{abstract}

\section{Introduction}

Support  vector  machines  (SVM)  are  now  routinely  used  for  many
classification problems  in machine learning~\cite{Sholkopf_Smola} due
to their  ease of use and  ability to generalize.  In  the basic case,
the input data,  corresponding to two groups, is  mapped into a higher
dimensional space,  where a  maximum-margin hyperplane is  computed to
separate  them.  The  ``kernel  trick''  is used  to  ensure that  the
mapping into higher dimensional  space is never explicitly calculated.
This can  be formulated as a non-negative  quadratic programming (NQP)
problem    and    there   are    efficient    algorithms   to    solve
it~\cite{platt98sequential}.

SVM  can be  trained using  variants  of the  gradient descent  method
applied   to  the   NQP.   Although   these  methods   can   be  quite
efficient~\cite{fcc-kafslps-98}, their drawback  is the requirement of
setting  the   learning  rate.    Subset  selection  methods   are  an
alternative      approach     to      solving     the      SVM     NQP
problem~\cite{platt98sequential}.   At  a  high  level  they  work  by
splitting the  arguments of the  quadratic function at  each iteration
into two sets: a fixed set, where the arguments are held constant, and
a  working  set  of  the  variables being  optimized  in  the  current
iteration.   These methods~\cite{platt98sequential},  though efficient
in space  and time,  still require a  heuristic to  exchange arguments
between the working and the fixed sets.

An alternative algorithm for solving  the general NQP problem has been
applied   to  SVM   in~\cite{FeiShaLee03}.    The  algorithm,   called
M\textsuperscript{3},  uses   multiplicative  updates  to  iteratively
converge to the solution. It  does not require any heuristics, such as
setting the learning rate or choosing how to split the argument set.

In this  paper we reformulate the  dual SVM problem  and demonstrate a
connection   to   the    non-negative   matrix   factorization   (NMF)
algorithm~\cite{citeulike:531251}.  NMF employs multiplicative updates
and is  very successful in practice  due to its  independence from the
learning rate parameter, low  computational complexity and the ease of
implementation.    The   new   formulation   allows   us   to   devise
multiplicative updates for solving  SVM with non-negative kernels (the
output value of the kernel function is greater or equal to zero).  The
requirement  of a non-negative  kernel is  not very  restrictive since
their set includes many  popular kernels, such as Gaussian, polynomial
of  even  degree  etc.   The  new  updates possess  all  of  the  good
properties   of  the   NMF  algorithm,   such  as   independence  from
hyper-parameters,  low  computational   complexity  and  the  ease  of
implementation.  Furthermore, the  new algorithm converges faster than
the   previous   multiplicative    solution   of   the   SVM   problem
from~\cite{FeiShaLee03} both asymptotically  (a proof is provided) and
in practice.   We also  show how  to solve the  SVM problem  with soft
margin using the new algorithm.

\section{NMF}
We present  a brief  introduction to NMF  mechanics with  the notation
that is  standard in NMF literature.  NMF  is a tool to  split a given
non-negative  data matrix into  a product  of two  non-negative matrix
factors~\cite{citeulike:531251}.   The  constraint  of  non-negativity
(all  elements  are  $\geq   0$)  usually  results  in  a  parts-based
representation  and is different  from other  factorization techniques
which result in more holistic representations (e.g.  PCA and VQ).

Given  a  non-negative  $m  \times  n$ matrix  $\bm{X}$,  we  want  to
represent  it with  a product  of two  non-negative  matrices $\bm{W},
\bm{H}$ of sizes $m \times r$ and $r \times n$ respectively:
\begin{equation}
\bm{X} \approx \bm{W} \bm{H}.
\end{equation}

Lee    and   Seung~\cite{citeulike:531251}    describe    two   simple
multiplicative updates  for $\bm{W}$ and  $\bm{H}$ which work  well in
practice.    These  correspond   to  two   different   cost  functions
representing the quality of approximation.  Here, we use the Frobenius
norm for the  cost function.  The cost function  and the corresponding
multiplicative updates are:
\begin{eqnarray}
E = \frac{1}{2}\|\bm{X} - \bm{W}\bm{H}\|_F \\\label{eqn:Wup}
\bm{W} = \bm{W}\odot \frac{\bm{X}\bm{H}^{T}}
				{\bm{W}\bm{H}\bm{H}^{T}}, \hskip 1cm
\bm{H} = \bm{H}\odot \frac{\bm{W}^{T}\bm{X}}
			{\bm{W}^{T}\bm{W}\bm{H}},
\end{eqnarray}
where $\|.\|_F$  denotes the Frobenius  norm and the  operator $\odot$
represents    element-wise   multiplication.     Division    is   also
element-wise.  
It should be noted that the cost function to be minimized is convex in
either $\bm{W}$  or $\bm{H}$ but  not in both~\cite{citeulike:531251}.
In~\cite{citeulike:531251}  it  is  proved  that  when  the  algorithm
iterates  using  the  updates~\eqref{eqn:Wup}, $\bm{W}$  and  $\bm{H}$
monotonically decrease the cost
function.  

\section{SVM as NMF}

Let  the set  of labeled  examples  $\{(\vec{x}_i,y_i)\}_{i=1}^N$ with
binary class labels  $y_i = \pm 1 $ correspond  to two classes denoted
by $A$ and $B$ respectively.  Let the mapping $\Phi(\vec{x}_i)$ be the
representation  of the  input datapoint  $\vec{x}_i$ in  space $\Phi$,
where  we  denote  the space  by  the  name  of the  mapping  function
performing  the  transformation.   We  now  consider  the  problem  of
computing the maximum margin hyperplane  for SVM in the case where the
classes  are  linearly separable  and  the  hyperplane passes  through
origin.

The dual quadratic  optimization problem for SVM~\cite{Sholkopf_Smola}
is given by minimizing the following loss function:
\begin{align}
  S(\vec{\alpha}) = \frac{1}{2} \sum_{i,j
    =  1}^{n}  \alpha_i  \alpha_j  y_i y_j  k(\vec{x}_i,\vec{x}_j)  -
  \sum_{i=1}^{n} \alpha_i\\
  \nonumber \mbox{subject to } \alpha_i \ge 0, i \in \{1..n\},
  \label{eq:svmdual}
\end{align}
where  $k(\vec{x}_i,\vec{x}_j)$ is  a kernel  that computes  the inner
product  $\Phi(\vec{x}_i)^{T}\Phi(\vec{x}_j)$ in  the space  $\Phi$ by
performing all operations only in the original data space on $x_i$ and
$x_j$, thus defining a Hilbert space $\Phi$.

The first sum can be split into three terms: two terms contain kernels
of elements  that belong  to the same  respective class (one  term per
class), and the third contains only the kernel between elements of the
two  classes.  This  rearrangement of  terms allows  us to  drop class
labels    $y_i,y_j$   from    the   objective    function.    Denoting
$k(\vec{x}_i,\vec{x}_j)$  with  $k_{ij}$  and  defining  $\rho_{ij}  =
\alpha_i \alpha_j k_{ij}$ for conciseness, we get:
\begin{align}
  \min_{\vec{\alpha}}   \frac{1}{2}  \left(\sum_{ij   \in  A}\rho_{ij}  - 2  \sum_{\begin{subarray}{c} i \in  B \\  j \in
        A\end{subarray}}\rho_{ij} +\sum_{ij \in B}\rho_{ij} \right) -\sum_{i=1}^{n}\alpha_i\\\nonumber
    \mbox{subject to  } \alpha_i \ge 0, i \in \{1..n\}.
  \end{align}
  Noticing    the   square    and    the   fact    that   $k_{ij}    =
  \Phi(\vec{x}_i)^{T}\Phi(\vec{x}_j)$ we rewrite the problem as:
\begin{align}
    \label{eq:nmf}
    \min_{\vec{\alpha}}         \frac{1}{2}
  \|\Phi(\bm{X}_{A})\vec{\alpha}_{A}
    - \Phi(\bm{X}_{B})\vec{\alpha}_{B}\|^2_2 - \sum_{i \in
      \{A,B\}}\alpha_i 
    \\\nonumber
    \mbox{subject to } \alpha_i \ge 0,
\end{align}
where  the matrices  $\bm{X}_{A}, \bm{X}_{B}$  contain  the datapoints
corresponding  to groups $A$  and $B$  respectively with  the stacking
being column-wise.  The map $\Phi$  applied to a matrix corresponds to
mapping each individual  column vector of the matrix  using $\Phi$ and
stacking   them   to   generate   the   new   matrix.    The   vectors
$\vec{\alpha}_{A},\vec{\alpha}_{B}$   contain   coefficients  of   the
support vectors of the two groups $A,B$ respectively.  We will use the
vector   $\vec{\alpha}$  to  denote   the  concatenation   of  vectors
$\vec{\alpha}_A,\vec{\alpha}_B$.   Expression~\eqref{eq:nmf} resembles
NMF with an  additional term in the objective~\cite{citeulike:531251}.
The       above      formulation      enables       other      metrics
$D(\Phi(\bm{X_{A}})\vec{\alpha}_A   ||  \Phi(\bm{X}_B)\vec{\alpha}_B)$
than   least  squares   for   SVM  such   as   more  general   Bregman
divergence~\cite{BregmanNMF}. However, to be computationally efficient
the metric used has to admit the use of the kernel trick.


\section{Multiplicative algorithm}
\label{sec:MUNK}

In this paper, we focus on kernel functions which are non-negative.  A
kernel function is non-negative when  its output value is greater than
or equal to  zero for all possible inputs in its  domain. We note that
quite  a  few of  the  commonly  used  kernels are  non-negative  like
Gaussian, polynomials of even degree,  etc.  We take the derivative of
the objective~\eqref{eq:nmf} with respect to $\vec{\alpha}_A$:
\begin{eqnarray} \nonumber
  \frac{\partial S}{\partial \vec{\alpha}_A}  &=&
  \Phi(\bm{X}_A)^{T}\Phi(\bm{X}_A) \vec{\alpha}_A 
  - \Phi(\bm{X}_A)^{T}\Phi(\bm{X}_B)\vec{\alpha}_B - \vec{1} \\\nonumber
  &=& 
  K(\bm{X}_A,\bm{X}_A) \vec{\alpha}_{A} - ( K(\bm{X}_{A},\bm{X}_{B})
  \vec{\alpha}_{B}+ \vec{1} ) 
\end{eqnarray}

We  slightly abuse  notation to  define  a matrix  kernel as  follows:
$K(\bm{C},\bm{D})$ is  given by the matrix  whose $(i,j)^{th}$ element
is given by  the inner product of $i^{th}$  and $j^{th}$ datapoints of
matrices   $\bm{C}$,$\bm{D}$  respectively   in   the  feature   space
$\bm{\Phi}$  for all values  of $(i,j)$  in range.   We note  that the
derivative  has a positive  and a  negative component.   Similarly, we
take the  derivative with respect to  $\vec{\alpha}_B$.  Recalling the
updates   for  NMF   from  previous   section,  we   write   down  the
multiplicative updates for this problem~\eqref{eq:nmf}:
\begin{eqnarray} \nonumber
  \vec{\alpha}_{A} & = & \vec{\alpha}_{A} \odot \frac{ K(\bm{X}_{A},\bm{X}_{B})
    \vec{\alpha}_{B}+ \vec{1} } 
  { K(\bm{X}_A,\bm{X}_A) \vec{\alpha}_{A} } \\ 
  \vec{\alpha}_{B} & = & \vec{\alpha}_{B} \odot \frac{ K(\bm{X}_{B},\bm{X}_{A}) 
    \vec{\alpha}_{A} + \vec{1}} 
  { K(\bm{X}_B,\bm{X}_B) \vec{\alpha}_{B} },
  \label{eq:upMUNK}
\end{eqnarray}
where $\vec{1}$ is  an appropriately sized vector of  ones and $\odot$
denotes  Hadamard  product as  before.   We  call  this new  algorithm
Multiplicative Updates for Non-negative Kernel SVM (\emph{MUNK}).

The  convergence  of  the  above  updates follow  from  the  proof  of
convergence           of           the           regular           NMF
updates~\cite{citeulike:531251}. Furthermore, since the Hessian of the
joint problem  of estimating $\vec{\alpha}_A$  and $\vec{\alpha}_B$ is
positive semi-definite  the alternating  updates have no  local minima
only the global minimum.

\section{Soft Margin}

We can extend the multiplicative updates to incorporate upper bound
constraints of the form $\alpha_i \le l$ where $l$ is 
a constant as follows: 
\begin{equation}
\alpha_i = \min{\{\alpha_i,l\}} 
\label{eq:cutoff}
\end{equation}
These are referred to as  box constraints, since they bound $\alpha_i$
from both above and below.

The dual problem for soft margin SVM is given by:
\begin{eqnarray}
\min_{\vec{\alpha}} S(\vec{\alpha}), \quad \mbox{subject  to }  0  \le \alpha_i  \le C, i \in \{1..n\},
  \label{eq:svmsoftdual}
\end{eqnarray}
The parameter  $C$ is a regularization  term, which provides  a way to
avoid overfitting.  Soft margin  SVM involves box constraints that can
be   handled  by   the   above  formulation.    At   each  update   of
$\vec{\alpha}$,  we  implement a  step  given by~\eqref{eq:cutoff}  to
ensure  the   box  constraint  is  satisfied.    This  corresponds  to
potentially reducing the step size  of the multiplicative update of an
element  and since  the problem  is convex  this will  still guarantee
monotonic decrease of the objective.

\section{Asymptotic convergence}

Sha et  al.~\cite{FeiShaLee03} observed  a rapid decay  of non-support
vector   coefficients  in   the  M\textsuperscript{3}   algorithm  and
performed  an analysis  of the  rate of  asymptotic  convergence. They
perturb    one    of     the    non-support    vector    coefficients,
e.g. $\vec{\alpha}_i$, away from the fixed point to some nonzero value
$\delta\vec{\alpha}_i$  and fix  all the  remaining  values.  Applying
their multiplicative  update gives a  bound on the asymptotic  rate of
convergence.

Let $d_i  = K(\vec{x}_i,\vec{w})/\sqrt{K(\vec{w},\vec{w})}$ denote the
perpendicular distance  in the feature  space from $\vec{x}_i$  to the
maximum  margin hyperplane and  $d =  \min_i d_i  = 1/\sqrt{K(\vec{w},
  \vec{w})}$  denote  the   one-sided  margin  to  the  maximum-margin
hyperplane.  Also,  $l_i = \sqrt{K(\vec{x}_i,\vec{x}_i)}$  denotes the
distance of  $\vec{x}_i$ to the origin  in the feature space  and $l =
\max_i l_i$ denote  the largest such distance. The  following bound on
the asymptotic rate of convergence $\gamma_i^{M^3}$ was established:

\begin{equation}
 \gamma_i^{M^3} \le \left[ 1 + \frac{1}{2}\frac{(d_i-d)d}{l_i l} \right]^{-1} 
\end{equation}

We perform  a similar analysis  for rate of asymptotic  convergence of
the multiplicative updates  of the MUNK algorithm.  We  perturb one of
the non-support vector coefficients  fixing all the other coefficients
and apply  the multiplicative update.  This enables us to  calculate a
bound  on rate  of  convergence. A  bound  on the  asymptotic rate  of
convergence in terms of geometric quantities is given as follows:

\begin{equation}
 \gamma_i^{MUNK} \le \left[ 1 + \frac{(d_i-d)d}{l_i l} \right]^{-1} 
\end{equation}

The proof sketch can be found  in appendix.  We note that our bound is
tighter  compared  to  the  M\textsuperscript{3} algorithm  as  $\gamma_i^{MUNK}  \le
\gamma_i^{M^3}$.

\section{Experiments}
\begin{table}
\begin{center}
\renewcommand{\arraystretch}{1.8}
\begin{tabular}{| l | l | c | c | c || c | c | c |}\hline
\multicolumn{2}{|c|}{\multirow{2}{*}{Kernel}} & \multicolumn{3}{|c||}{Breast} &
\multicolumn{3}{|c|}{Sonar}\\ \cline{3-8}
\multicolumn{2}{|c|}{} & M\textsuperscript{3} & M & KA &
M\textsuperscript{3} & M & KA \\ \hline
\multirow{2}{4mm}{\begin{sideways}{Poly}\end{sideways}}&
4 &2.26 &2.26 &2.26 & 9.62&9.62 &9.62\\\cline{2-8}
&6&3.76 &3.76 &3.76 &10.58&10.58&10.58\\\hline
\multirow{2}{4mm}{\begin{sideways}{Gaussian}\end{sideways}} &
3 & 2.26&2.26 &2.26 & 11.53&11.53 &11.53 \\\cline{2-8}
&1& 0.75 &0.75 &0.75 &7.69&7.69 &7.69 \\\hline
\end{tabular}
\end{center}
\caption{Misclassification rates (\%) on the breast cancer and sonar
  datasets after convergence of the M\textsuperscript{3}, MUNK (M) and
  Kernel Adatron (KA) algorithms. Polynomial kernels of degree 4 and 6
  and Gaussian kernels of $\sigma$ 1 and 3 were used.}
\label{tbl:rates}
\end{table}
In order to demonstrate the practical applicability of the theoretical
properties proved  in previous section,  we test the above  updates on
two  real  world problems  consisting  of  breast  cancer dataset  and
aspect-angle     dependent    sonar     signals    from     the    UCI
Repository~\cite{Newman+Hettich+Blake+Merz:1998}.   They  contain  683
and 208 labelled examples respectively.  The breast cancer dataset was
split into 80\% and 20\% for training and test sets respectively.  The
sonar dataset  was equally divided  into training and test  sets.  The
vectors $\vec{\alpha}$  were initialized  the same in  all algorithms.
Different kernels involving polynomial and radial basis functions were
applied to  the dataset.  For  comparison we also provide  results for
the M\textsuperscript{3} and Kernel-Adatron (KA)~\cite{fcc-kafslps-98}
algorithms.  Misclassification rates on the test datasets are shown in
Table~\ref{tbl:rates}.  They match  previously reported error rates on
this dataset~\cite{FeiShaLee03}.

These  results  support  our  derivations  and  demonstrate  that  the
algorithm  can be  used for  training SVM  with  non-negative kernels.
However,  since  the problem  is  convex  and  there exists  a  unique
solution all correct algorithms will converge to the same solution and
arrive at the same classification error rates.

\begin{figure}
\centering
\includegraphics[width=\linewidth]{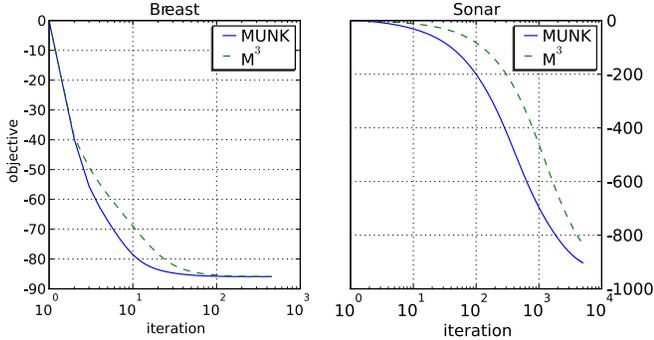}
\caption{Convergence  of the objective  with iterations  when training
  with  Gaussian kernel  ($\sigma  = 3$).   Lower  curve means  faster
  convergence.  Note  that  $x$  axis  is  logarithmic,  indicating  a
  multiplicative speedup for MUNK over a wide operating range. }
\label{fig:speed}
\end{figure}

MUNK is slightly faster per iteration than M\textsuperscript{3} due to
an extra  square root and  multiplication per training pattern  in the
M\textsuperscript{3} algorithm.  We  ignore that slight difference and
plot   the   objective   function    per   iteration   of   MUNK   and
M\textsuperscript{3}  algorithms  on  the  Breast and  Sonar  sets  in
Figure~\ref{fig:speed}.   The  result  agrees with  the  theoretically
shown   upper  bound:   MUNK  converges   about  twice   as   fast  as
M\textsuperscript{3}.

\section{Conclusions}

We  have derived simple  multiplicative update  rules for  solving the
maximum-margin classifier  problem in SVMs  with non-negative kernels.
No  additional parameter  tuning is  required and  the  convergence is
guaranteed.   The  updates  are  straight-forward  to  implement.   
The updates could also be used  as part of a subset method which could
potentially  speed up  MUNK  algorithm.  MUNK  shares  the utility  of
M\textsuperscript{3}  algorithm in  that it  is easy  to  implement in
higher-level languages like MATLAB with application to small datasets.
It also  shares the drawback of M\textsuperscript{3}  in its inability
to directly  set a variable to  zero.  However, we have  shown MUNK to
have  an  asymptotically  faster   rate  of  convergence  compared  to
M\textsuperscript{3}  algorithm   and  we  believe   this  provides  a
motivation for further research  in multiplicative updates for support
vector machines.   Also the derivation  was constructed in such  a way
that  it highlights  the connection  between SVM  with  a non-negative
kernel  and NMF.   Since  multiplicative updates  emerge in  different
settings and algorithms it might be interesting to find the pattern of
when such updates  are possible and how to  automatically derive them.
Our presentation  of NMF  and SVM correspondence  can be  considered a
step towards this direction.


\bibliographystyle{IEEEbib}
\bibliography{papers}

\section*{Appendix}
Let      the     fixed      point     be      $\vec{\alpha}^*$     and
$K(\bm{X}_A,\bm{X}_A)\vec{\alpha}_A^*$  be denoted by  $\vec{z}^+$ and
$K(\bm{X}_A,\bm{X}_B)\vec{\alpha}_B^*$  by $\vec{z}^-$.  If  we choose
an $i$th non-support vector coefficient from $\vec{\alpha}_A$, then we
have $\vec{z}_i^+  -\vec{z}_i^- \ge 1$. Let  the multiplicative factor
be denoted by $\gamma_i$.  We then have:
\begin{align}\nonumber
  \frac{1}{\gamma_i} = \frac{\vec{z}_{i}^{+}}{\vec{z}_i^- + 1}
  = 1 + \frac{\vec{z}_{i}^+ - \vec{z}_{i}^- -1}{\vec{z}_{i}^- + 1}
  &\ge& 1 +  \frac{K(\vec{x}_i,\vec{w}) - 1}{\vec{z}_{i}^+}  
\end{align}
where $\vec{w}  = \sum_i \alpha^*_i x_i  y_i$ is the  normal vector to
the maximum margin hyperplane.  We have used the following:

\begin{align}\nonumber
\vec{z}_i^+ -\vec{z}_i^- =\sum_{j\in A} k_{ij}\vec{\alpha}_j^*
			- \sum_{k\in B} k_{ik}\vec{\alpha}_k^*
			= K(\vec{x}_i,\vec{w}),
\end{align}
where $k_{ij} = K(\vec{x}_i,\vec{x}_j)$.

We now obtain a bound on the denominator:
\begin{align}\nonumber
\vec{z}_i^+ = \sum_{j \in A} K(\vec{x}_i,\vec{x}_j) \vec{\alpha}_j^* 
            \le \max_{k \in A} K(\vec{x}_i,\vec{x}_k) 
	    \sum_{j\in A} \vec{\alpha}_j^* \\\nonumber
	    \le \sqrt{K(\vec{x}_i,\vec{x}_i)} 
	    \max_{k \in A} \sqrt{K(\vec{x}_k,\vec{x}_k)} K(\vec{w},\vec{w}) 
\end{align}
We have used  the Cauchy-Schwartz inequality for kernels  and an upper
bound for the sum of vector $\alpha_A^*$.

We do  a similar  analysis by perturbing  an $i$th  non-support vector
coefficient from group B. Combining the analysis, the lower bound is:
\begin{equation*}
\frac{1}{\gamma_i} \ge 1 + \frac{ K(\vec{x}_i,\vec{w}) - 1} { \sqrt{K(\vec{x}_i,
		\vec{x}_i)} \max_k \sqrt{K(\vec{x}_k,\vec{x}_k)} K(\vec{w},
		\vec{w})}
\end{equation*}

\end{document}